%% file: main.tex
\pdfoutput=1
\documentclass[letterpaper,twocolumn,10pt]{article}
\PassOptionsToPackage{hyphens}{url}
\usepackage{usenix2019_v3}
\usepackage{amsmath}
\usepackage{amssymb}
\usepackage{booktabs}
\usepackage{multirow}
\usepackage{xcolor}
\usepackage{graphicx}
\usepackage{stfloats}

\usepackage{etoolbox}
\makeatletter
\patchcmd{\@maketitle}{\vbox to 2.5in}{\vbox to 1.35in}{}{\PackageError{layout}{Title patch failed}{}}
\makeatother

\hypersetup{hypertexnames=false}
\newcommand{\sys}{FrontierReplay}
\newcommand{\ms}{\,ms}
\newcommand{\fig}{figures}

\begin{document}

\date{}

\title{\Large \bf FrontierReplay: Split-Free Exact Prefix Caching for Hybrid Language Models}

\author{\rm Zekun Wang \quad Yizhuo Wang$^{*}$\\[2pt]
\textit{School of Computer Science, Beijing Institute of Technology}}

\maketitle
\begingroup
\renewcommand{\thefootnote}{\fnsymbol{footnote}}
\footnotetext[1]{Corresponding author.}
\endgroup

% ======================================================================
\begin{abstract}
Hybrid language models combine self-attention with recurrent layers whose in-place state updates restrict prefix reuse to exact saved positions. Saving states on a coarse grid wastes reusable prefixes, whereas existing fine-grained exact caches must cut the prefill forward pass at each mid-prompt checkpoint, adding scheduler steps that outweigh the reuse gains under load.

We observe that across multi-turn chat, agent loops, document question answering, and tree search, prefix reuse almost always ends at structural boundaries predictable in advance, such as turn ends and known branch points. We present \mbox{\textbf{\sys{}}}, a prefix caching system that materializes exact checkpoints at these boundaries \emph{without} cutting the prefill. While the forward pass runs uncut, FrontierReplay logs a short window of recurrent-layer inputs and invokes a cross-layer, state-only replay kernel to build the target state 6--37$\times$ cheaper per token than full recomputation without host synchronization. A frontier-aware retention policy prioritizes active anchors and proven branches while lending dead checkpoint slots to the attention KV cache.

Across diverse hybrid architectures, model scales, and workloads, FrontierReplay delivers clear gains over existing exact caches, lowering median time to first token (TTFT) by 16--89\% (up to 8.9$\times$), raising 80\% service-level objective (SLO) capacity by up to 1.73$\times$, and improving goodput by up to 4.26$\times$ (3.67$\times$ at larger scales). Compared with approximate replay methods, FrontierReplay delivers up to 1.54$\times$ higher goodput while preserving exact task accuracy, avoiding the up to 17\% accuracy drop suffered by approximation on long contexts.
\end{abstract}

% ======================================================================
\section{Introduction}
\label{sec:intro}

\begin{figure}[t]
\centering
\includegraphics[width=0.92\columnwidth]{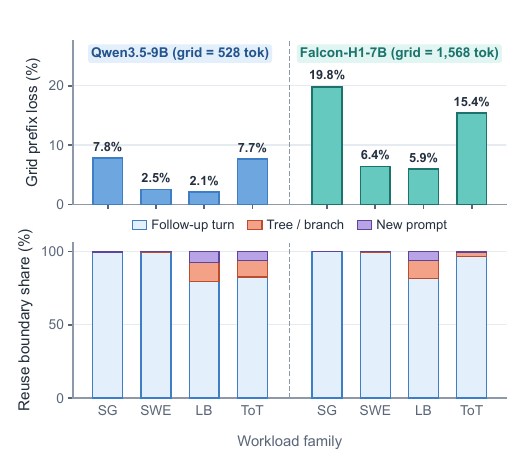}
\vspace{-4pt}
\caption{\textbf{Reuse ends at predictable positions that vLLM's state grid misses.} Measured across each workload trace with an unbounded KV cache. (a)~Share of reusable prefix tokens that lie past the last state block boundary of vLLM. (b)~Share of reusable prefix tokens by where the prefix ends. A turn boundary is the template stable end or the end of a finished turn. A known branch point is one where two earlier requests have already diverged.}
\label{fig:reuse}
\end{figure}

Hybrid language models combine self-attention with recurrent state-space or linear-attention layers. Qwen3.5, Falcon-H1, Nemotron-3, Jamba, Kimi-K3, and GLM-5.3-Flash interleave the two layer types or run them in parallel inside each block, using Mamba-2, Gated DeltaNet (GDN), or Kimi Delta Attention as the recurrent component~\cite{qwen35,falconh1,nemotron3,jamba,kimik3,glm53flash,mamba2,gdn,kda}. Their recurrent layers keep a fixed-size state instead of a key-value (KV) entry per token, which is what makes long contexts cheap. The same property makes prefix caching hard. Serving systems avoid recomputing a shared prefix by reusing cached KV pages~\cite{vllm,sglang,promptcache,preble}, and attention KV can be truncated at any page boundary. A recurrent state cannot: it folds the whole history into one tensor that is overwritten as each token is processed. A cache can reuse it only at the exact positions where a copy was saved.

vLLM, a leading open-source serving engine, saves these copies on a coarse grid. With prefix caching enabled, its default \emph{align} mode cuts each prefill chunk at the next multiple of the recurrent state block (528 tokens for Qwen3.5-9B, 1568 for Falcon-H1-7B) and keeps the state at each grid point~\cite{vllm030}. A request that shares a prefix up to some position restores the last grid point before it and recomputes the rest through the full model. Figure~\ref{fig:reuse}(a) shows the cost on four public workloads: the grid cannot serve 2--8\% of the reusable prefix tokens on Qwen and 6--20\% on Falcon, and those tokens are recomputed by every attention, feed-forward (MLP), and recurrent layer. Equally, when the single tail checkpoint of a request in align mode lands in request-unique tokens, or is recycled before the next turn arrives, the next turn misses the whole prompt~\cite{vllm030}.

Finer-grained exact caches exist. Marconi keeps states at request ends and at branch points of a radix tree~\cite{marconi}; Sparse Prefix Caching (SPC) places a few checkpoints per prompt where a learned overlap distribution says hits will land~\cite{sparseprefix}. Both still face the mechanism behind the grid: a forward pass exposes the recurrent state only where it ends, so a checkpoint in the middle of a prompt requires ending a forward pass there. Each such cut adds a scheduler step and a short trailing chunk to the request. Under load this is not free. With identical checkpoint placement and identical prefill tokens, cutting at each target takes 1.44--1.86$\times$ as many prefill steps as our split-free design on chat, agent, and tree-search traces (\S\ref{sec:mot-cut}), and SPC runs 1.44--1.52$\times$ as many prefill steps as vLLM on tree search and chat. Approximate caches such as SuffixReplay~\cite{suffixreplay} and Tail-Replay~\cite{tailreplay} sidestep the problem by rebuilding an approximate state from the last few hundred tokens at hit time, at the cost of changing the model's outputs.

Our starting observation is that reuse positions in these workloads are not arbitrary. Figure~\ref{fig:reuse}(b) classifies where each reused prefix ends. On every workload and both models, 79--100\% of reusable tokens end at an earlier turn's boundary: the position where the chat template's generation prompt begins, or the end of a finished turn. Most of the rest ends at a branch point the server has already seen, such as a shared system prompt or a document that earlier questions diverged from. The server can therefore name the position of the next hit before the next request arrives. What it lacks is a cheap way to have an exact state there.

\sys{} supplies that state with \emph{split-free checkpointing}. The prefill forward pass for a request runs uncut. While it runs, the runtime logs the recurrent-layer inputs over the short window between the request's starting state and the target boundary; afterwards a cross-layer, state-only replay kernel recomputes only the recurrent updates over that window and writes an exact checkpoint at the target. Replay skips attention, MLPs, and output projections, so it costs 6--37$\times$ less per token than recomputing the span (Table~\ref{tab:r9-microbench}) and adds only 1--3\ms{} of GPU work per checkpoint on an A800 without host synchronization. Unlike concurrent in-kernel side-output export (FlashStates~\cite{flashstates}), which modifies native kernels and couples cost to operator chunk tiles, post-forward replay leaves forward kernels untouched and decouples 16-token boundaries from backend tile size (\S\ref{sec:eval-ablation}). Every cache hit is then a plain state copy. Because each checkpoint is large (49--67\,MiB) and a cache can hold only a few hundred, \sys{} pairs split-free production with a retention policy built around the conversation frontier: checkpoints whose KV is gone are dead, ancestors superseded by a deeper checkpoint on the same path are dropped next, speculative ones follow, while proven branch points and the anchors of running requests are kept. Checkpoint slots that hold only dead state are lent to the attention KV pool.

We build \sys{} on vLLM~0.30.0 and evaluate it on Qwen3.5-9B (interleaved GDN) and Falcon-H1-7B (parallel Mamba-2) across ShareGPT conversations, SWE-agent tool sessions, LongBench document questions, and GSM8K tree search, against vLLM, Marconi, SPC, and SuffixReplay at a 32\,GiB cache budget. Across all eight operating anchors, \sys{} is the best exact cache (1.11--4.26$\times$ vLLM goodput), raises 80\% SLO capacity by up to 1.73$\times$, and cuts median TTFT by 16--89\% (1.1\,s to 125\ms{} on Falcon SWE-agent). On tensor-parallel 27B and 34B models, it cuts median TTFT by 1.4--7.7$\times$ across all workloads and delivers up to 3.67$\times$ goodput on 34B tree search. Against approximate SuffixReplay, it ties on chat and agent traces and wins by 1.10--1.54$\times$ at high load while preserving exact long-context accuracy (where SuffixReplay drops 13--17\% at 24K--32K tokens).

In summary, this paper makes four contributions:
\begin{itemize}
  \item A measurement of where prefix reuse lands in hybrid-model serving and what producing an exact state there costs: reuse ends at predictable positions off the state grid, and cutting the prefill to reach them adds prefill steps that outweigh the reuse under load (\S\ref{sec:motivation}).
  \item Split-free checkpointing, which produces exact recurrent checkpoints at 16-token boundaries within a bounded 1024-token replay window—without cutting the forward pass or modifying native kernels—using a short input log and tile-agnostic state-only replay (\S\ref{sec:splitfree}--\S\ref{sec:planner}).
  \item Frontier-aware checkpoint management and serving integration: a boundary planner that needs no future trace information, tiered retention with long-prefix protection, and lending of dead checkpoint slots to the KV pool (\S\ref{sec:planner}--\S\ref{sec:serving}).
  \item An implementation on vLLM and an evaluation on two hybrid architectures at five model sizes from 4B to 34B, four workloads, and three exact and one approximate baseline, including task-accuracy tests (\S\ref{sec:impl}--\S\ref{sec:eval}).
\end{itemize}

% ======================================================================
\section{Background}
\label{sec:background}

\paragraph{LLM inference.} A large language model (LLM) generates text autoregressively, and serving a request proceeds in two phases~\cite{orca,sarathi,distserve}. In the \emph{prefill} phase the engine processes all prompt tokens in parallel and produces the first output token. Prefill is compute-bound, and its duration, together with any time the request waits in queue, determines the \emph{time to first token} (TTFT). In the \emph{decode} phase the engine then produces one token per forward pass, each conditioned on everything before it. Decode is bound by memory bandwidth, and its pace determines the \emph{time per output token} (TPOT). So that decode does not recompute the past at every step, each attention layer keeps the keys and values of earlier tokens in a key-value (KV) cache.

Serving engines batch at iteration granularity: every scheduler step runs one forward pass over a batch that combines decode tokens of running requests with prefill tokens of newly admitted ones~\cite{orca,vllm}. \emph{Chunked prefill} splits a long prompt into chunks bounded by a per-step token budget, so that a long prefill does not stall the decode of other requests~\cite{sarathi}. \emph{Paged attention} stores the KV cache in fixed-size pages, which avoids fragmentation and lets requests share pages~\cite{vllm}, and engines replay each step's kernels from pre-captured CUDA graphs to keep launch overhead low~\cite{vllm,sglang,flashinfer,flashattn2,cudagraphs}.

\paragraph{Prefix caching.} Many requests repeat a long prefix of an earlier one: the history of a multi-turn conversation, the system prompt and tool outputs of an agent, or the document of a document-question workload. Prefix caching keeps the KV pages of finished requests and indexes 16-token pages by chained content hashes, so a new request reuses the longest cached page prefix and runs prefill only for the rest~\cite{vllm,sglang,promptcache,cachedattention}. Each reused token saves prefill computation, which shortens TTFT and leaves more of every batch to other requests. \sys{} works in this part of the system: it decides at which positions exact recurrent states are kept while a prompt is prefilled, produces them without changing how the prefill is chunked, and copies them back when a later request hits the cache. The model's decode kernels are unchanged, and the state at the end of each finished turn is also kept.

\paragraph{Hybrid models.} Self-attention gives direct access to every earlier token, but its KV cache grows with context length~\cite{vaswani2017,vllm}. State-space models (SSMs) and linear-attention layers instead fold the past into a fixed-size state~\cite{linearattn,rwkv,mamba1,mamba2,gla}. Because a fixed-size state limits exact recall, current model families combine both, either interleaving recurrent and attention layers (Qwen3.5 uses three GDN layers per attention layer) or running recurrent and attention heads in parallel inside each block (Falcon-H1)~\cite{qwen35,falconh1,waleffe24}.

\paragraph{Recurrent state updates.} A recurrent layer reads tokens in order and keeps, for each head, a state matrix. In Mamba-2 the state after token $t$ is
\begin{equation}
S_t=\alpha_t S_{t-1}+v_t k_t^{\top},\qquad o_t=S_t q_t,
\label{eq:mamba}
\end{equation}
where $S_t\in\mathbb{R}^{d_v\times d_k}$ and $d_k$ and $d_v$ are the key and value dimensions of the head. The key $k_t\in\mathbb{R}^{d_k}$, value $v_t\in\mathbb{R}^{d_v}$, and query $q_t\in\mathbb{R}^{d_k}$ are projections of the layer's input at token $t$ (called $B_t$, $x_t$, and $C_t$ in the Mamba-2 paper~\cite{mamba2}); $\alpha_t\in(0,1)$ is a decay computed from the same input, and $o_t$ is the head's output at token $t$. GDN adds a delta-rule correction~\cite{deltanet,gdn},
\[
S_t=\alpha_t S_{t-1}\bigl(I-\beta_t k_t k_t^{\top}\bigr)+\beta_t v_t k_t^{\top},
\]
in which $\beta_t\in(0,1)$ is a per-token write strength and $I$ is the identity matrix: before writing the new key--value association, the layer erases what the state stored along $k_t$. In both architectures a short causal convolution over the last few inputs precedes the recurrence, and its sliding window is part of the layer's state. Unrolling Eq.~\eqref{eq:mamba} over the $h$ tokens that follow position $c$ gives
\begin{equation}
S_{c+h}=\gamma_{1:h}\,S_c+\sum_{j=1}^{h}\gamma_{j+1:h}\,v_j k_j^{\top},\qquad \gamma_{i:j}=\prod_{\ell=i}^{j}\alpha_\ell ,
\label{eq:unroll}
\end{equation}
where $j$ indexes the tokens after $c$ (so $v_j$ belongs to token $c+j$), $\gamma_{i:j}$ is the decay accumulated from the $i$-th through the $j$-th of them, and the empty product $\gamma_{h+1:h}$ equals 1. GDN admits the same form once the product of its $(I-\beta_j k_j k_j^{\top})$ factors is resolved chunk by chunk in WY representation, a compact form for products of rank-one updates~\cite{wy1987,gdn,replayssm}. The state at any later position is therefore a function of an earlier state and of the recurrent-layer inputs $(\alpha_j,\beta_j,k_j,v_j)$ in between; attention, MLPs, and the outputs $o_t$ are not needed to compute it. \sys{} relies on this identity to produce states after the fact.

\begin{figure}[t]
\centering
\includegraphics[width=0.95\columnwidth]{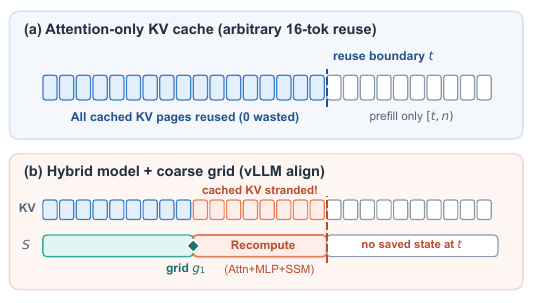}
\vspace{-4pt}
\caption{\textbf{Granularity mismatch in hybrid prefix caching.} (a)~Attention-only models reuse 16-token KV pages up to any reuse boundary $t$. (b)~With in-place recurrent state $S$ saved only on a coarse grid $g_1$, cached KV pages in $[g_1,t)$ cannot be resumed alone, forcing full recomputation over that span.}
\label{fig:bg-mismatch}
\end{figure}

\paragraph{Prefix caching for hybrid models.} Attention KV can be reused page by page, but a recurrent layer must resume from its state at exactly the reuse position, so the reusable prefix of a hybrid model ends at the deepest position whose recurrent state was saved (Figure~\ref{fig:bg-mismatch}). vLLM's align mode saves states at multiples of a state block, whose size is chosen so that one state occupies as many bytes as the attention pages of that block~\cite{vllm030}; an opt-in mode that also saves the state at a shared-prefix junction requires speculative decoding.

% ======================================================================
\section{Characterization and Design Opportunities}
\label{sec:motivation}

A fine-grained exact cache has to answer two questions: at which positions should it keep recurrent states, and how does it obtain a state at a position that a forward pass does not end at? We study both on the four workloads of our evaluation (\S\ref{sec:eval-setup}), starting with where reuse actually ends.

\subsection{Where does reuse end?}
\label{sec:mot-where}

Across each workload trace, Figure~\ref{fig:reuse} records the longest prefix each request shares with any earlier sequence. Two observations follow.

\emph{A coarse grid wastes a visible share of reusable work.} The last grid point before the reuse boundary leaves 2--8\% of reusable tokens unserved on Qwen (528-token blocks) and 6--20\% on Falcon (1568-token blocks); ShareGPT and tree search lose the most, since much of their reuse is short relative to a block.

\emph{Reuse boundaries are structural.} In a multi-turn conversation, the next turn re-renders the history and the shared prefix stops where the chat template closes the previous turn or begins its generation prompt. Agent sessions behave the same way, and their first calls additionally share a long system prompt. In LongBench, questions about one document diverge where the document ends; in tree search, siblings diverge at step boundaries. Across all eight model--workload pairs, 79--100\% of reusable tokens end at an earlier turn's boundary, and at most 8\% end at a branch point that no earlier request has revealed. The position of most future hits is therefore known when the previous request is served. Knowing where to keep a state, however, is only half of the problem; the cache must also obtain that state.

\subsection{Producing a state there requires cutting the prefill}
\label{sec:mot-cut}

\begin{figure}[t]
\centering
\includegraphics[width=\columnwidth]{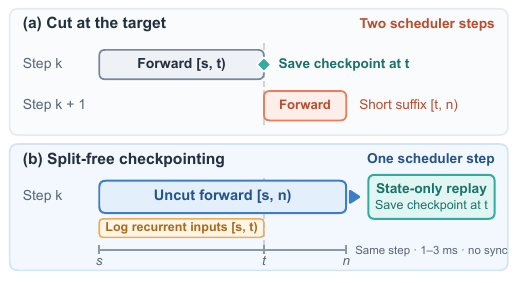}
\caption{\textbf{Two ways to obtain the exact state at a reuse boundary $t$.} (a)~Ending the forward at $t$ saves the checkpoint in step $k$ and leaves the short suffix $[t,n)$ for step $k+1$. (b)~\sys{} runs one uncut forward over $[s,n)$ and logs recurrent-layer inputs for $[s,t)$. State-only replay then writes the checkpoint at $t$ in the same scheduler step without host synchronization.}
\label{fig:splitfree}
\end{figure}

A prefill forward pass leaves only one recurrent state behind: the state after its last token. To keep a state at a boundary $t$ inside a prompt, an exact cache must end a forward pass at $t$ and run the remaining tokens as a separate, usually short, chunk in the next scheduler step (Figure~\ref{fig:splitfree}a). vLLM's align mode does exactly this at grid points, Marconi does it at branch points, and SPC at each placed checkpoint.

The price of a cut is an extra prefill step, not extra tokens. To measure it in isolation, consider two configurations of our system that keep checkpoints at the same boundaries and therefore compute the same prefill tokens, one cutting the forward pass at each boundary and one not. Cutting inflates prefill steps by 1.44--1.86$\times$ on ShareGPT, SWE-agent, and ToT, and 1.05--1.09$\times$ on LongBench where cuts are sparse (Figure~\ref{fig:mot-cost}a). SPC shows the same effect: on Qwen tree search it computes fewer prefill tokens than vLLM yet runs 1.44$\times$ as many prefill steps, and 1.52$\times$ as many on Falcon ShareGPT. Every extra step delays the request's first token and occupies a batch slot that decode steps of other requests could use. Under load, the steps cost more than the extra reuse saves. On Falcon SWE-agent, the cutting configuration reaches 1.92$\times$ the goodput of vLLM, while the same checkpoints obtained without cuts reach 3.47$\times$ (\S\ref{sec:eval-ablation}). Table~\ref{tab:space} summarizes how each design obtains a state at a mid-prompt boundary, and the next subsection shows that a cut is not the only way.

\subsection{State-only replay is cheap}
\label{sec:mot-replay}

\begin{figure*}[t]
\centering
\includegraphics[width=0.92\textwidth]{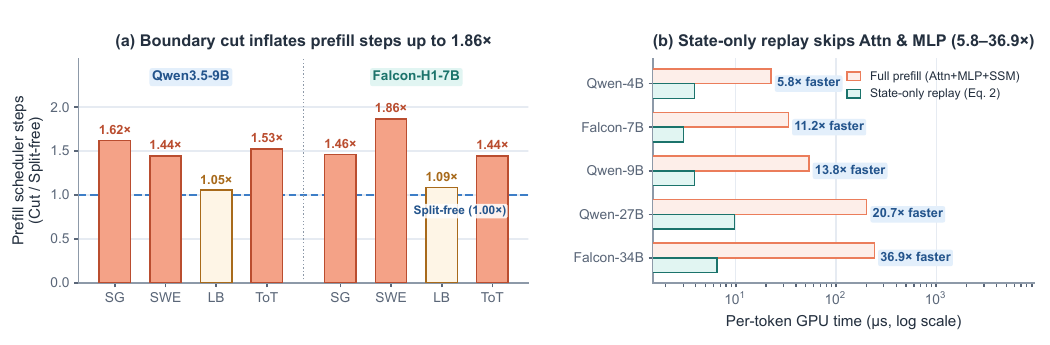}
\vspace{-4pt}
\caption{\textbf{Scheduler-step penalty of boundary cuts versus GPU efficiency of state-only replay.} (a)~Cutting the forward pass at each reuse boundary inflates prefill scheduler steps by 1.05--1.86$\times$ relative to split-free execution with identical checkpoints and token counts. (b)~Per-token GPU replay latency $b$ versus full prefill slope $c$ across five hybrid models (Table~\ref{tab:r9-microbench}); skipping attention and MLPs yields a 5.8--36.9$\times$ leverage that widens with model scale.}
\label{fig:mot-cost}
\end{figure*}

Equation~\eqref{eq:unroll} suggests another option: let the forward pass run uncut from the chunk start $s$, keep the recurrent-layer inputs for the window $[s,t)$, and recompute only the recurrence afterwards. Figure~\ref{fig:mot-cost} contrasts the scheduler-step cost of boundary cuts with the GPU cost of state-only replay, and Table~\ref{tab:r9-microbench} gives the fitted model parameters. We fit the GPU time of replaying a span of $L$ tokens through all recurrent layers as $a+b\cdot L$, where the intercept $a$ is the fixed cost of one replay (kernel launches and reading the starting state) and the slope $b$ is the cost per replayed token. Likewise, $c$ denotes the per-token slope of an ordinary prefill over the same span. The intercept $a$ is 1.0--3.8\ms{}, and $b$ is 5.8--36.9$\times$ smaller than $c$ (Figure~\ref{fig:mot-cost}b), because replay executes no attention, MLP, normalization, or output projection. The ratio $c/b$ grows with model size (5.8$\times$ for Qwen3.5-4B, 20.7$\times$ for 27B, 36.9$\times$ for Falcon-H1-34B), since the recurrent state grows more slowly than the rest of the model. Comparing replayed states against those of an uninterrupted prefill over the same tokens, we find them identical bit for bit, so replay can stand in for the forward pass without changing results.

\input{tables/tab1-microbench.tex}

\subsection{Checkpoints are large and few}
\label{sec:mot-few}

Cheap production raises the next question: how many checkpoints can a cache afford to keep? A checkpoint holds the recurrent and convolution state of every recurrent layer: 49.1\,MiB for Qwen3.5-9B (and 4B) and 66.9\,MiB for Falcon-H1-7B. Within a 32\,GiB cache budget, a checkpoint pool holds only 105--208 of them, while 32 concurrent requests each occupy working states and every turn of every session creates new candidates. Which ones to keep matters more than in a KV cache. A frequency- or recency-based policy keeps a conversation's older ancestors, which every later turn passes through, and evicts the newest anchor, which is the only one the next turn can use. Speculative checkpoints, such as the end of each prompt or of each finished turn, crowd out branch points that have already proven useful. A good policy must therefore know where each checkpoint stands relative to the sessions that may still reach it.

\begin{table}[t]
\centering
\caption{\textbf{How exact and approximate caches obtain recurrent state at a reuse boundary.}}
\label{tab:space}
\footnotesize
\setlength{\tabcolsep}{3pt}
\begin{tabular}{@{}llcc@{}}
\toprule
Method & State at a mid-prompt boundary & Cut & Exact \\
\midrule
vLLM align~\cite{vllm030} & forward ends at the block grid & yes & yes \\
Marconi~\cite{marconi} & forward ends at a branch point & yes & yes \\
SPC~\cite{sparseprefix} & forward ends at a planned point & yes & yes \\
SuffixReplay~\cite{suffixreplay} & approximate rebuild at every hit & no & no \\
\textbf{\sys{}} & \textbf{state-only replay after the forward} & \textbf{no} & \textbf{yes} \\
\bottomrule
\end{tabular}
\end{table}

% ======================================================================
\section{System Design}
\label{sec:design}

\begin{figure*}[t]
\centering
\includegraphics[width=0.95\textwidth]{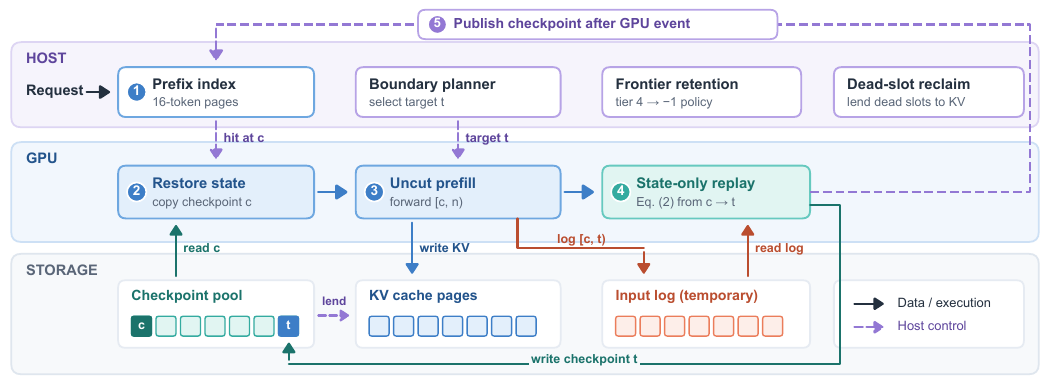}
\vspace{-4pt}
\caption{\textbf{\sys{} architecture.} The host scheduler maintains a prefix index over 16-token pages and manages checkpoint placement, retention, and dead-slot reclaim. \textcircled{\scriptsize 1}~The index returns the deepest checkpoint $c$ on the cached prefix. \textcircled{\scriptsize 2}~The worker copies it into the request's state slot. \textcircled{\scriptsize 3}~The uncut prefill over $[c,n)$ writes KV pages and stages recurrent-layer inputs for $[c,t)$ in a temporary log. \textcircled{\scriptsize 4}~State-only replay starts from the saved state at $c$ and writes the checkpoint at $t$ into a free slot. \textcircled{\scriptsize 5}~The scheduler publishes the checkpoint after the GPU event completes. Solid arrows show data and execution flow. Dashed arrows show control flow, including the reassignment of dead checkpoint slots to the KV pool.}
\label{fig:arch}
\end{figure*}

\subsection{Overview}
\label{sec:overview}

\sys{} decouples \emph{when} checkpoints are materialized from \emph{how} prefill chunks are scheduled, separating host control from GPU state execution (Figure~\ref{fig:arch}). On the host, the scheduler tracks a radix tree over 16-token chained prefix hashes, each checkpoint's frontier tier, and the pool mapping between checkpoints and KV blocks. On the GPU, HBM is partitioned into three disjoint regions managed over stable virtual addresses: the paged attention KV cache, a checkpoint pool backed by CUDA virtual memory (VMM), and a bounded staging log of 16-token input pages.

Each request passes through five stages:
(1)~\emph{Lookup and restore}: When a request arrives, the radix index locates the longest resident KV prefix $[0,b)$ and the deepest resident checkpoint $c \le b$. The scheduler leases $c$ so it cannot be evicted while in use and schedules a state copy into the request's working slot, leaving $[c,n)$ to run under ordinary chunked prefill.
(2)~\emph{Boundary planning} (\S\ref{sec:planner}): Using only structural cues visible at admission—chat-template ends, finished turns, and known branch junctions—the planner selects mid-chunk targets $t_1 < \dots < t_m$ within a 1024-token replay window (capping wider spans to proactive cuts).
(3)~\emph{Uncut forward and logging} (\S\ref{sec:splitfree}): The prefill step executes without cuts; entry-point hooks copy each recurrent layer's inputs over $[s,t_m)$ and precision-rounding bitmasks into request-private log pages.
(4)~\emph{Asynchronous replay and publication} (\S\ref{sec:splitfree}, \S\ref{sec:serving}): On the same CUDA stream, a cross-layer replay kernel advances the leased state over the logged inputs to materialize each checkpoint without host synchronization, and the host exposes new checkpoints to lookups and frees the log once the step's GPU event retires.
(5)~\emph{Frontier retention and reclaim} (\S\ref{sec:retention}, \S\ref{sec:reclaim}): A tiered eviction policy preserves running anchors and proven branches, while a periodic controller reassigns excess dead checkpoint pages to expand the KV pool without breaking CUDA graphs.

\subsection{Split-free checkpointing}
\label{sec:splitfree}

When the planner assigns targets $t_1<\dots<t_m$ inside a prefill chunk that starts at $s$, \sys{} keeps the chunk intact and does three things.

\paragraph{Lease the starting state.} Replay needs the exact state at $s$, and the forward pass overwrites the working state. The scheduler therefore requires a resident checkpoint at $s$: the one the request restored from, the natural checkpoint saved at the end of its previous chunk, or the empty state at position~0. It leases that checkpoint so it cannot be evicted before the replay runs, and reserves a free slot for the output. If no such checkpoint exists, the target falls back to a cut.

\paragraph{Log the window.} During the forward pass, the inputs of every recurrent layer for positions $[s,t_m)$ are copied into 16-token log pages, storing the per-token projections, gate inputs, and any required page-end convolution state (Table~\ref{tab:r9-microbench}). Log pages are private to the request, so concurrent siblings that share a prefix never contend for them, and they are invisible to lookups.

\paragraph{Replay state-only.} Right after the forward pass, in the same scheduler step, a replay kernel applies Eq.~\eqref{eq:unroll} to every recurrent layer, starting from the leased state and running over the logged window; it writes the state at each target into the reserved checkpoint slot. The kernel is designed to be both cheap and exact:
(a)~it computes only state transitions and never the per-token outputs $o_t$;
(b)~it reaches all recurrent layers through device-side pointer tables, so one replay takes a handful of kernel launches rather than one launch sequence per layer; and
(c)~it reproduces the numerics of the forward pass that produced the log. Each log page records, as bitmasks, the chunk and decode-step boundaries at which that pass rounded the state to its storage precision, and replay splits the span and rounds at the same positions.
Replay is enqueued on the same CUDA stream as the forward pass and as every later reader of the checkpoint, so stream order alone guarantees that the inputs are ready and the result is complete before use; the scheduler never waits for the GPU. The checkpoint is published to the index once the GPU event recorded after the replay completes, and the log pages are then released.

The window is bounded by design. Before dispatch, targets $>1024$ tokens past chunk start are capped to a proactive cut; at runtime, \textsf{prepare} degrades safely to an exact cut if no live start anchor or free log page is available. Because the log is staging memory rather than a persistent cache, in-flight windows remain compact (median $96$, p90 $768$ tokens; $99.5\%$ within-window completion at 32\,GiB). Compared with in-kernel export (\S\ref{sec:eval-ablation}), post-forward replay trades this bounded staging footprint (412--630\,MiB replay workspace plus a 1.14--1.39\,GiB VMM input-log slab, totaling 1.76--1.79\,GiB within a 2\,GiB budget) for zero edits to native kernels or CUDA-graph signatures, dynamic target selection, and latency decoupled from coarse tiles ($C{=}256$).

\subsection{Boundary planner}
\label{sec:planner}

The planner chooses targets using only information that is available when a request is scheduled.
\begin{itemize}
  \item \emph{Template stable end.} At startup \sys{} renders the model's chat template on two probe conversations and compares the generation prompt of one turn with how the next turn re-renders that turn. The common prefix defines the model's trailing template suffix. For every request, the position just before this suffix becomes a target, so the next turn finds an exact anchor at its reuse boundary.
  \item \emph{End of a finished turn.} When decoding ends, the state at the last 16-token boundary of prompt plus output is copied into a checkpoint; it costs no replay because decode states pass through every page boundary.
  \item \emph{Branch junction.} If a request's KV match ends at a radix node that already has two children and no checkpoint, and the node is at least 256 tokens deep, the node becomes a target for this request, so every later sibling restores exactly there. System prompts of agent sessions and documents shared by several questions are found this way.
  \item \emph{Chunk ends.} Chunked prefill ends every chunk on a 16-token boundary; the state there is available for free and is offered to the pool as a low-priority \emph{natural} checkpoint.
\end{itemize}
For Falcon-H1 the planner never places or restores a state below 1568 tokens. Short Mamba-2 states stored in bfloat16 (BF16) are sensitive to rounding before the decay has damped the initial history.

\subsection{Frontier retention}
\label{sec:retention}

Every resident checkpoint belongs to one tier, and eviction removes the checkpoint with the lowest tier, breaking ties by least recent use:
\begin{itemize}
  \item \emph{Dead} ($-1$): some KV page of its prefix has been evicted, so no lookup can reach it.
  \item \emph{Superseded} (0): a deeper resident checkpoint lies on the same path with no branch junction in between, and some other request has already continued past it; or it is a natural checkpoint and a deeper natural checkpoint exists on the same path.
  \item \emph{Speculative} (1--2): natural checkpoints (1) and template or finished-turn anchors not yet used (2).
  \item \emph{Proven} (3): a branch junction, or a checkpoint that at least two distinct requests have restored from.
  \item \emph{Running} (4): the anchor of a request that is still executing; its session's next turn will need it.
\end{itemize}
When the pool is full, a new natural checkpoint is admitted only if it would evict a dead or superseded one.

The ordering follows from how sessions move. A conversation's newest anchor is the only checkpoint its next turn can use, but under a frequency-based utility it looks least valuable, since no request has restored from it yet; evicting it while its producer is still decoding breaks the session's reuse chain, which is why running anchors rank highest. Branch junctions serve every future sibling, whereas the speculative anchor at a prompt end serves at most one follow-up, so junctions are protected from the stream of speculative anchors that each new prompt brings. Finally, promotion to the proven tier requires two \emph{distinct} restoring requests: in a multi-turn session every turn restores once from its predecessor, so a single restore would make every anchor of every conversation permanent.

\paragraph{Retaining long prefixes.}
The checkpoint at the end of a document query can include question-specific tokens and lie beyond the prefix shared with later queries. Under eviction pressure, \sys{} protects one existing natural checkpoint per long-prefix group: the deepest chunk boundary within one prefill budget (2048 tokens) before the template-stable end and at least two budgets (4096 tokens) from the start. While its KV remains resident and no deeper reachable junction exists, this checkpoint receives the same priority as a proven checkpoint (tier~3). Recent lookups also estimate how many reusable tokens would be lost without each checkpoint; within each tier, eviction removes lower-loss checkpoints first and promotes repeatedly hit checkpoints whose estimated loss reaches one prefill budget.

\subsection{Dead-checkpoint reclaim}
\label{sec:reclaim}

The checkpoint and KV pools have separate budgets, and pressure can be lopsided. Under client-side queueing, a session may wait between calls while other requests churn the KV pool; its prompt-end KV pages are evicted, its anchor becomes dead, and the checkpoint pool fills with dead states while the KV pool is the bottleneck. \sys{} therefore runs a controller every 16 requests. It counts checkpoints that died because their KV was evicted; if no checkpoint above tier 0 has been evicted in the last four periods (the checkpoint pool is not binding), it moves up to $\min(16, 2d)$ slots, where $d$ is the dead count, from the checkpoint pool to the KV pool, evicting only tier $\le 0$ checkpoints and never shrinking the pool below a floor that covers the working states of running requests. Memory moves at the granularity of CUDA virtual-memory pages~\cite{cudavmm}, so both pools keep stable addresses for CUDA graphs~\cite{cudagraphs,vattention}. Its own evictions are not counted as signal, so the controller cannot feed itself.

\subsection{Serving integration}
\label{sec:serving}

\paragraph{Decode.} \sys{} tracks 16-token page states during decode for turn-end reuse, and inserts at most one isolated decode step when one-token decodes coincide with pending prefills, leaving decode kernels unchanged.

\paragraph{Execution.} Replay, copy, and log kernels are compiled and captured during engine warmup, before any request is served. Pages and checkpoints become visible to lookups only after the GPU event of the step that wrote them, so a preempted or failed step never exposes partial state. All decisions above run in the scheduler process at request granularity.

% ======================================================================
\section{Implementation}
\label{sec:impl}

\sys{} is built on vLLM~0.30.0~\cite{vllm,vllm030}. It replaces vLLM's prefix cache for hybrid models and keeps the rest of the engine, including model execution, attention kernels, the paged KV cache, and CUDA graphs. It hooks the scheduler where prefill chunks are formed, the recurrent-state block manager, and the prefill and decode entry points of the recurrent layers, and it adds the replay kernels (Triton~\cite{triton}), the radix index and policy (Python), and a pool manager over CUDA virtual memory~\cite{cudavmm}. GDN and Mamba-2 need different log formats and replay kernels (\S\ref{sec:splitfree}) but share all policy code. The design requires only a hash-indexed paged KV cache, control over prefill chunk ends, and access to recurrent-layer inputs. For comparison we implemented the baselines on the same runtime (\S\ref{sec:eval-setup}). Marconi's admission and eviction policy is ported to the serving path and checked against the authors' reference implementation~\cite{marconi}, and SPC follows its paper~\cite{sparseprefix}, including the dynamic program that places its checkpoints. The authors of SuffixReplay have not released their source code, so we implemented it independently from the description in their paper~\cite{suffixreplay}. All exact baselines share our runtime, vLLM's decode path, and the same restore floor, so differences between them come from where and how they keep state.

% ======================================================================
\section{Evaluation}
\label{sec:eval}

Our evaluation answers four questions: (1)~Does \sys{} improve end-to-end goodput, latency, and SLO capacity over vLLM and existing exact caches (\S\ref{sec:eval-main})? (2)~How does it compare against approximate replay in throughput and task accuracy (\S\ref{sec:eval-approx})? (3)~How much does each design component contribute, how does it behave under tighter cache budgets, and what overhead does it add (\S\ref{sec:eval-ablation})? (4)~How do these gains scale across model sizes (\S\ref{sec:eval-scale})?

\subsection{Setup}
\label{sec:eval-setup}

\paragraph{Platform, models, and default scope.} Unless specified otherwise, all runs use one NVIDIA A800 80\,GB GPU with vLLM~0.30.0, BF16 weights, chunked prefill (2048-token budget), piecewise CUDA graphs, and a 32\,GiB cache budget under a joint SLO of TTFT $\le 500$\ms{} and mean TPOT $\le 50$\ms{}. The primary models are Qwen3.5-9B~\cite{qwen35} (24 GDN and 8 attention layers, FP32 recurrent state, 528-token block) and Falcon-H1-7B~\cite{falconh1} (44 parallel hybrid blocks, BF16 Mamba-2 state, 1568-token block). Qwen3.5-4B, Qwen3.5-27B, and Falcon-H1-34B appear in \S\ref{sec:eval-scale}; the 27B and 34B models run with two-way tensor parallelism and fixed per-GPU pools.

\paragraph{Workloads and metrics.} We evaluate four public traces tokenized with each model's chat template: \emph{ShareGPT}~\cite{sharegpt} (4--8-turn windows, final prompt $\ge 4$K tokens), \emph{SWE-agent}~\cite{sweagent,swesmith} (4--8-call tool sessions, initial prompt $\ge 8$K tokens), \emph{LongBench}~\cite{longbench} (32 documents of 8--30K tokens with 195 Zipf(1.1) queries), and \emph{ToT}~\cite{gsm8k,tot,mathshepherd} (eight-branch GSM8K search trees). Sessions arrive as Poisson processes at offered load $\lambda$ (calls/s) with child calls issued upon parent completion and recorded responses replayed identically across systems. Every point runs in a cold process after untimed warmup; goodput divides SLO-passing calls (within a 180\,s timeout) by total wall time. Throughout \S\ref{sec:eval}, anchor, ablation, and multi-run 27B/34B points report means over three paired runs with 95\% Student-$t$ confidence intervals (shown as error bars or bracketed intervals), while full load/SLO sweeps, capacity budgets below 32\,GiB (except 16\,GiB Falcon ShareGPT), 4B runs, and single-point follow-ups (marked $^{\dagger}$) use one paired run per operating point.

\paragraph{Baselines.} \emph{vLLM} (align mode) uses its native 32\,GiB unified KV-and-state cache. The checkpoint-based exact baselines---\emph{Marconi} (request-end/branch states with FLOP-aware eviction~\cite{marconi}), \emph{SPC} (four DP-planned checkpoints per prompt~\cite{sparseprefix}), and \emph{Fixed-1024} (periodic 1024-token cuts)---receive 20\,GiB for KV and 12\,GiB for states. \sys{} partitions the 32\,GiB budget into 20\,GiB KV, 10\,GiB checkpoints, and 2\,GiB log/replay workspace; all exact methods share Falcon's 1568-token restore floor (\S\ref{sec:planner}). Approximate \emph{SuffixReplay}$^{*}$~\cite{suffixreplay} caches four hidden-state rows per 64-token page and rebuilds recurrent state from zero over $\min(5\%\cdot\text{hit},512)$ rows on interleaved models (Qwen).

\subsection{End-to-End Goodput, Latency, and Capacity}
\label{sec:eval-main}

\begin{figure*}[!t]
\centering
\includegraphics[width=0.95\textwidth]{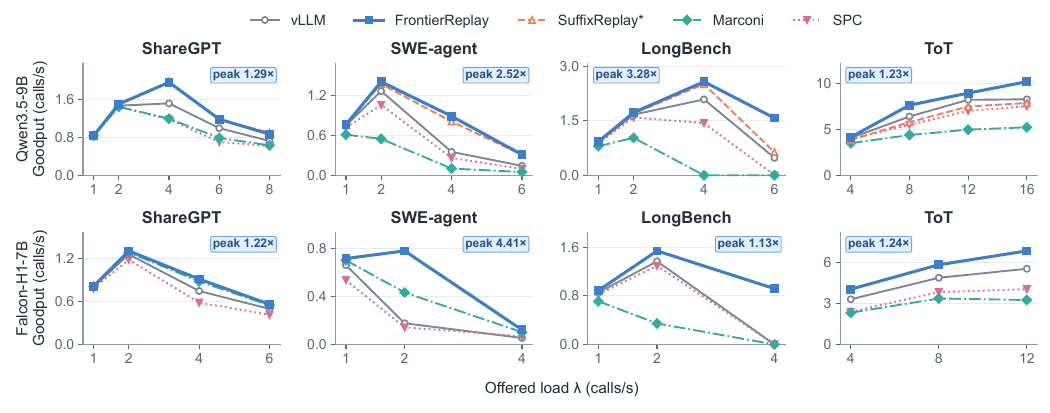}
\vspace{-4pt}
\caption{\textbf{Goodput versus offered load across both models and all four workloads.} SuffixReplay$^{*}$ (Qwen only) approximates the recurrent state and is evaluated alongside exact baselines under the default 32\,GiB budget.}
\label{fig:load}
\end{figure*}

Figure~\ref{fig:load} plots goodput across offered loads, and Table~\ref{tab:r10-summary} consolidates the anchor comparisons, key tail/median latencies, and 80\% SLO capacities. \sys{} outperforms every exact cache across all eight model--workload pairs. Gains are largest where structural reuse ends between vLLM's state grid points and saved recomputation keeps the server out of overload: on Falcon SWE-agent ($\lambda{=}2$), where vLLM recomputes up to 1567 tokens per agent call and stalls into overload, \sys{} reaches 4.26$\times$ vLLM's goodput and cuts median TTFT from 1.1\,s to 125\ms{} (eliminating request failures that make vLLM's p99 latency unbounded). On Qwen ShareGPT and LongBench, \sys{} improves anchor goodput by 1.27$\times$ and 1.23$\times$, widening to 3.28$\times$ at $\lambda{=}6$ on Qwen LongBench as vLLM collapses under load. Marconi and SPC trail vLLM at most anchors because Marconi's request-end checkpoints trail chat-template boundaries and SPC's periodic checkpoints cut prefill mid-prompt.

\input{tables/tab3-main-summary.tex}

On workloads where vLLM already captures most document or prefix blocks, frontier retention unlocks the remaining bottleneck. On Falcon LongBench ($\lambda{=}2$), protecting one deep natural checkpoint per long-prefix group leaves only 4 requests with $\ge 8192$ tokens of resident KV lacking a recoverable state, cuts scheduled prefill tokens by 33.6\%, lifts token hit rate from 0.88 to 0.92, and raises anchor goodput to 1.13$\times$; at $\lambda{=}4$, where all other exact methods drop to zero SLO attainment, \sys{} still achieves 22.8\% SLO attainment (0.92 calls/s, median TTFT 10.45\,s). On two-token ToT search steps ($\lambda{=}8$), \sys{} keeps Qwen's TPOT violations at zero (p99 TPOT 41 vs.\ 75\ms{}, 481 vs.\ 427 passing calls, 1.13$\times$ goodput, extending to 1.23$\times$ at $\lambda{=}16$) while holding median TTFT at 68\ms{} (below vLLM's 80\ms{}). Across all eight anchors, \sys{} achieves the lowest median TTFT (reduced by 16--46\% on Qwen and 24--89\% on Falcon) and the lowest p99 TPOT among exact methods.

\begin{figure*}[!tb]
\centering
\includegraphics[width=0.95\textwidth]{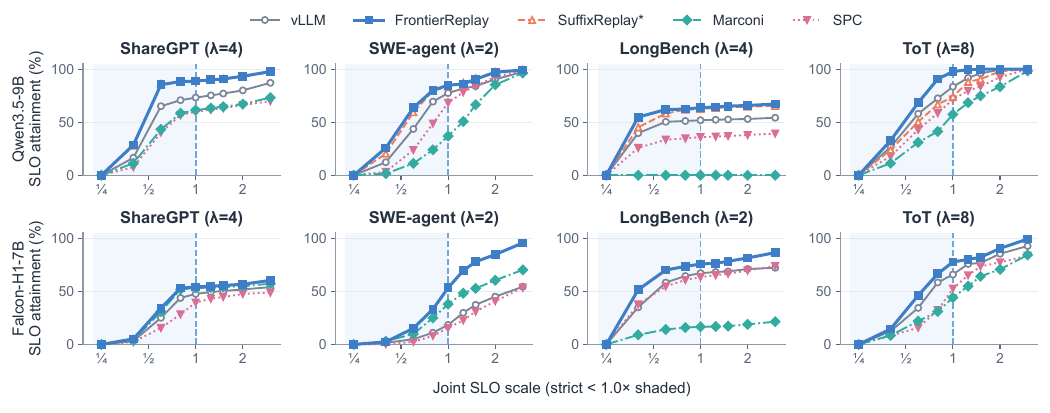}
\vspace{-4pt}
\caption{\textbf{SLO attainment under scaled latency targets} (scale~1 = 500\ms{} TTFT, 50\ms{} TPOT) at the anchor loads.}
\label{fig:sloscale}
\end{figure*}

Figure~\ref{fig:sloscale} varies SLO strictness from $0.25\times$ to $3\times$ the default bounds at the anchor loads. \sys{} consistently stays on top, with the widest margins under strict latency targets (e.g., 85\% vs.\ 65\% on Qwen ShareGPT and 64\% vs.\ 45\% on Qwen SWE-agent at $0.6\times$ SLO). Summarizing the full load sweep via 80\% SLO capacity (Table~\ref{tab:r10-summary}, right; using 80\% as in prior TPOT-constrained studies~\cite{distserve,sarathi,splitwise} since several workloads never reach 90\% even at minimum load), \sys{} increases sustainable load on all eight pairs by 1.15--1.33$\times$ on Qwen and 1.05--1.73$\times$ on Falcon (including a paired $\lambda{=}0.5$ sweep for Falcon SWE-agent).

\subsection{Exactness vs.\ Approximate Reuse}
\label{sec:eval-approx}

\input{tables/tab4-contested.tex}

\begin{figure}[!htb]
\centering
\includegraphics[width=\columnwidth]{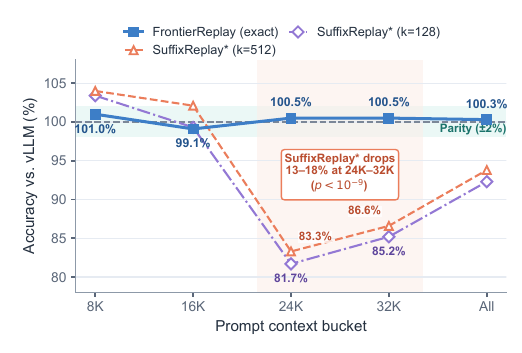}
\caption{\textbf{Exact replay preserves long-context accuracy; approximate reuse degrades at 24K--32K tokens.} RULER variable tracking on Qwen3.5-9B relative to vLLM.}
\label{fig:vt}
\end{figure}

SuffixReplay$^{*}$ avoids prefill cuts without checkpoints, making it the closest competitor on Qwen in Figure~\ref{fig:load}. In side-by-side runs at the highest contested loads (Table~\ref{tab:r9-contested}), \sys{} matches SuffixReplay$^{*}$ on ShareGPT and SWE-agent and wins on ToT at $\lambda{=}16$ (1.10$\times$) and LongBench at $\lambda{=}6$ (1.54$\times$): SuffixReplay$^{*}$ re-runs every recurrent block (including MLPs) over up to 512 rows on every hit, whereas \sys{} pays a 0.2\ms{} state copy at hit time and replays once per checkpoint.

More importantly, approximation alters outputs on long contexts. Over 303 LongBench, needle-in-a-haystack, and RULER variable-tracking (VT) items~\cite{longbench,ruler} plus an 800-question VT sweep (Figure~\ref{fig:vt}), greedy generation with \sys{} matches vLLM at 100.2--100.3\% relative score ($p \ge 0.34$ across all lengths, 92\% token-identical outputs, with the remaining differences arising solely from near-tie floating-point non-associativity across dynamic concurrent batch shapes---matching the logit deviation of other exact caches with zero non-tie argmax flips). By contrast, SuffixReplay$^{*}$ shifts outputs systematically (only 33\% token-identical) and loses 13--17\% relative accuracy on VT at 24K and 32K tokens for $k{=}512$ (15--18\% for $k{=}128$; $p<10^{-9}$).

\subsection{Mechanism Attribution, Memory Pressure, and Overhead}
\label{sec:eval-ablation}

\begin{figure}[!t]
\centering
\includegraphics[width=\columnwidth]{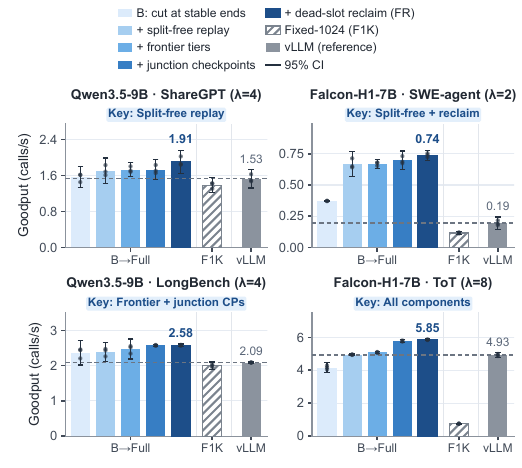}
\vspace{-6pt}
\caption{\textbf{Ablation on four representative bottlenecks.} Cumulative components (left to right) versus Fixed-1024 (periodic cuts) and vLLM (dashed line).}
\label{fig:ablation}
\end{figure}

\paragraph{Mechanism attribution.} Following our taxonomy, Figure~\ref{fig:ablation} breaks down four bottlenecks from variant \textsf{B} (cut-based stable anchors). \emph{Split-free checkpointing} yields the largest gain: removing cuts lifts Qwen ShareGPT (1.56 to 1.71 calls/s), raises Falcon SWE-agent (0.37 to 0.67 vs.\ 0.19 for vLLM), and cuts Falcon ToT median TPOT from 39 to 33\ms{} (4.17 to 4.95 calls/s). \emph{Frontier tiers and junction checkpoints} preserve shared prefixes on Qwen LongBench (2.39 to 2.58 calls/s; median TTFT 113 to 81\ms{}) and lift Falcon ToT to 5.77 calls/s (and Falcon LongBench from 0.90 to 1.13$\times$). \emph{Dead-checkpoint reclaim} frees stale slots under queueing, bringing Falcon SWE-agent to 0.74, Falcon ToT to 5.85 (1.19$\times$ vLLM), and Qwen ShareGPT to 1.91 calls/s (1.25$\times$). Across all 12 evaluated ablation settings (eight 32\,GiB anchors and four 16\,GiB Chat/Agent settings), \sys{} beats vLLM in goodput on every setting (with 95\% CIs strictly above 1.0$\times$) and lowers median TTFT on 11 of 12, while Fixed-1024 trails vLLM everywhere.

\begin{figure}[!tb]
\centering
\includegraphics[width=\columnwidth]{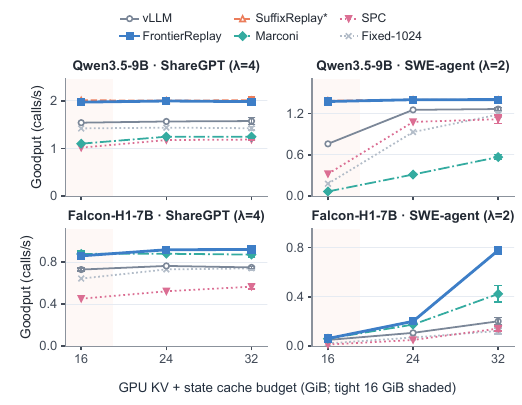}
\vspace{-6pt}
\caption{\textbf{Goodput across tighter cache budgets (16, 24, and 32\,GiB).} Evaluates ShareGPT ($\lambda{=}4$) and SWE-agent ($\lambda{=}2$) as available GPU KV and state memory is halved down to 16\,GiB.}
\label{fig:capacity}
\end{figure}

\paragraph{Tighter cache budgets.} Figure~\ref{fig:capacity} halves the budget down to 16\,GiB on ShareGPT and SWE-agent. \sys{} leads all exact baselines at 11 of 12 points: on Qwen SWE-agent at 16\,GiB, elastic VMM borrowing preserves $98.7\%$ split-free completion ($97$ log drops) and 1.37 versus 0.76 calls/s for vLLM (1.81$\times$), while Marconi and SPC collapse to 0.07 and 0.32 calls/s. On Falcon SWE-agent at 16--24\,GiB, all systems saturate below 0.21 calls/s as tighter checkpoint slots ($66$--$82$)—not log exhaustion ($0$ drops)—trigger $7.2\%$ fallback to exact cuts.

\paragraph{System overhead.} Across all eight 32\,GiB anchors, restores ($\approx 0.2$\ms{} median, $<0.9$\ms{} p99) and replay (1.2--3.1\ms{}) consume $\le 1.2\%$ of GPU wall time, and host scheduling adds $\le 1.2\%$. The $1.76$\,GiB (Qwen: $0.62{+}1.14$) and $1.79$\,GiB (Falcon: $0.40{+}1.39$) allocations are static provisioned workspace-plus-VMM-record slabs; mean active writes are only $68.4$ and $23.9$\,MiB/step ($47.0$\,MiB overall), sustaining $99.5\%$ within-window completion ($0.5\%$ runtime fallback across 30 runs) while cutting prefill tokens by 12--77\% over vLLM.

\begin{figure}[!tb]
\centering
\includegraphics[width=\columnwidth]{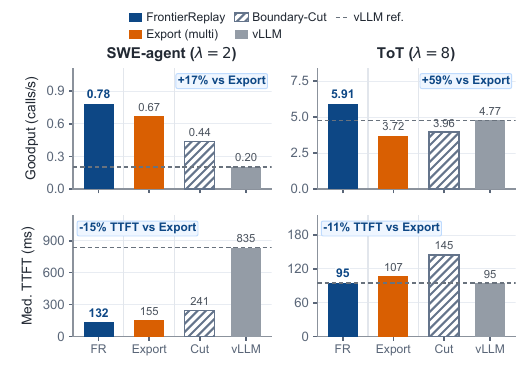}
\vspace{-6pt}
\caption{\textbf{Post-forward replay vs.\ in-kernel export on Falcon-H1-7B$^{\dagger}$.} Off-tile cuts trigger partial-tile scans in Mamba-2 ($C_{\text{tile}}{=}256$), letting \sys{} outperform multi-target export, cutting, and vLLM on SWE-agent and ToT.}
\label{fig:falcon-export}
\end{figure}

\paragraph{In-kernel state export.$^{\dagger}$}
We compare \sys{} against multi-target in-kernel state export ($M_{\text{cap}}{=}8$, inspired by FlashStates~\cite{flashstates}) implemented on the same vLLM runtime. In-kernel export patches 3 native kernels (538\,LOC) and adds 3 CUDA-graph signatures versus zero for \sys{}, yielding a clear tile-width crossover. On fine-chunk GDN (Qwen3.5-9B, $C_{\text{tile}}{=}64$), export handles $m{\le}8$ targets in $O(1)$ time (19.7--19.9 vs.\ 23.2--36.0\ms{} isolated), edging out \sys{} by 1.5--2.5\% on ShareGPT and SWE-agent (2.02 vs.\ 1.97 and 1.40 vs.\ 1.38 calls/s). On coarse-chunk Mamba-2 (Falcon-H1-7B, $C_{\text{tile}}{=}256$), off-tile boundaries ($d\bmod 256\neq 0$) force partial-tile scans (+18.4--26.6\ms{} over native). Across Falcon workloads, \sys{} achieves +17\% goodput and 15\% lower median TTFT on SWE-agent (Figure~\ref{fig:falcon-export}; 0.78 vs.\ 0.67 calls/s, 132 vs.\ 155\ms{}), +59\% goodput and 11\% lower TTFT on ToT (5.91 vs.\ 3.72 calls/s, 95 vs.\ 107\ms{}), and +16\% goodput with 30\% lower TTFT on LongBench (1.56 vs.\ 1.34 calls/s, 82 vs.\ 116\ms{}).

\subsection{Scaling Across Model Sizes}
\label{sec:eval-scale}

\begin{figure*}[!t]
\centering
\includegraphics[width=0.88\textwidth]{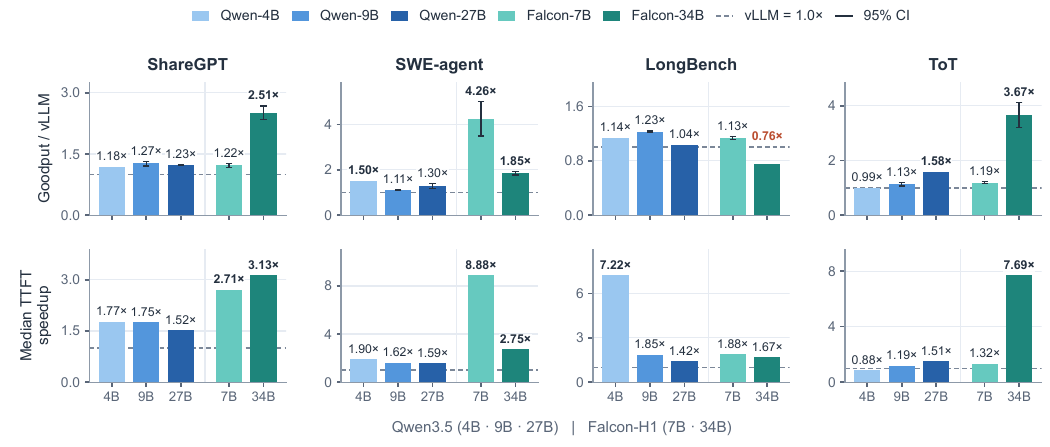}
\vspace{-5pt}
\caption{\textbf{Model scaling across five sizes (4B to 34B).} Loads are calibrated so vLLM stays below 1\,s median TTFT.}
\label{fig:scale}
\end{figure*}

Figure~\ref{fig:scale} evaluates Qwen3.5-4B (single GPU) and tensor-parallel Qwen3.5-27B and Falcon-H1-34B (two GPUs, 32\,GiB/GPU, calibrated so vLLM median TTFT stays below 1\,s). Because per-token staging records scale linearly $O(d)$ with width (363 to 957\,KiB/tok on Qwen; 312 to 729\,KiB/tok on Falcon, sharded across TP ranks with $0$ log drops) while prefill compute scales quadratically $O(d^2)$, replay leverage rises from 5.8$\times$ (4B) to 20.7$\times$ (27B) and 36.9$\times$ (34B) (Table~\ref{tab:r9-microbench}). \sys{} cuts median TTFT by 1.42--7.69$\times$ on every 27B and 34B workload and raises goodput on seven of eight: on 34B ToT, vLLM drops to 0.58 hit rate and 143\ms{} TPOT, whereas \sys{} holds 0.96 coverage for 3.67$\times$ goodput and 7.69$\times$ lower median TTFT; on 27B ToT, clearing decode TPOT contention (p99 168 to 67\ms{}) raises goodput to 1.58$\times$.

Two regimes mark the design boundary. On small 4B ToT ($\lambda{=}12$), which already has TPOT slack, eliminating all 18 TPOT misses trades +7\ms{} of median TTFT (55 to 62\ms{}) for 0.99$\times$ goodput (445 vs.\ 444 passing calls). On 34B LongBench ($\lambda{=}0.5$), \sys{} cuts median TTFT (225 to 135\ms{}) yet trails vLLM in goodput (0.76$\times$, 0.83 vs.\ 0.90 hit rate): static tensor-parallel sub-pools (22\,GiB KV + 82 state slots/GPU vs.\ vLLM's 32\,GiB unified cache) underperform a unified pool when workloads are KV-capacity-bound with sparse replay reuse and live tail slots block contiguous reclaim (22 blocked attempts), motivating dynamic cross-pool KV scaling as future work.

% ======================================================================
\section{Related Work}
\label{sec:related}

\paragraph{Exact prefix caching for hybrid models.} vLLM caches recurrent states on a block grid by cutting prefill chunks at grid points and, with speculative decoding, at detected junctions~\cite{vllm030}. SGLang stores states at step ends in a radix tree with separate LRU lists~\cite{sglang}. Marconi admits states at request ends and branch points, evicting by FLOP efficiency per byte~\cite{marconi}. SPC places fixed checkpoint counts per prompt via dynamic programming over an overlap histogram~\cite{sparseprefix}. All four obtain mid-prompt states by ending a forward pass there. Concurrent work FlashStates~\cite{flashstates} avoids cuts via in-kernel single-target export and dependency-aware eviction. \sys{} keeps structural checkpoints without touching native kernels (0 vs.\ 538 modified LOC), outperforming export on coarse-tile Mamba-2 (\S\ref{sec:eval-ablation}), and manages pools via frontier tiers and VMM dead-slot reclaim. HiState~\cite{histate} offloads checkpoints to SSD and DASC~\cite{dasc} compresses states; both are orthogonal to split-free replay.

\paragraph{Approximate reuse.} SuffixReplay rebuilds recurrent state from zero over sparse cached hidden states~\cite{suffixreplay}, Tail-Replay replays a trailing window~\cite{tailreplay}, and HybridCache reuses quantized or stale states~\cite{hybridcache}; all avoid cuts but alter long-context outputs.

\paragraph{Recurrent-state replay within a request.} ReplaySSM buffers recent inputs with a base state to cut decode writes and enable speculative rollback~\cite{replayssm}. \sys{} instead applies unrolled recurrence across requests to materialize prefix checkpoints rather than serving a single sequence.

\paragraph{Serving systems.} Orca, vLLM, SGLang, Llumnix, Sarathi-Serve, FlashInfer, FlashAttention-2, DistServe, Splitwise, LoongServe, and Mooncake established modern scheduling, paged KV, fast attention, and disaggregation~\cite{orca,vllm,sglang,llumnix,sarathi,flashinfer,flashattn2,distserve,splitwise,loongserve,mooncake}; CachedAttention, CacheBlend, PromptCache, and Preble share KV~\cite{cachedattention,cacheblend,promptcache,preble}, while vAttention manages VMM memory~\cite{vattention}. Our retention adapts GDSF~\cite{gdsf} with frontier position.

% ======================================================================
\section{Conclusion}

Hybrid models reuse prefixes only at exact recurrent states, and cutting prefill costs more than it saves under load. Because reuse ends at predictable template and branch boundaries, \sys{} builds exact checkpoints after an uncut forward pass via an input log and state-only replay kernel under frontier-aware retention. Across two architectures and four workloads, it leads every exact cache at all anchors (up to 4.26$\times$ vLLM goodput; 3.67$\times$ on 34B) while matching approximate speed with exact outputs.

\bibliographystyle{unsrt}
\bibliography{main}

\end{document}

%% file: tables/tab1-microbench.tex
\begin{table}[t]
\centering
\small
\caption{\textbf{State-only replay costs 6--37$\times$ less per token than recomputing the span through the model.} All-layer replay measured in isolation on one A800. The GPU time to replay a span of $L$ tokens is fitted as $a+b\cdot L$, where $a$ is the fixed cost of one replay and $b$ the cost per replayed token; $c$ is the per-token cost of an ordinary prefill over the same span, and leverage is $c/b$. Checkpoint is the size of one all-layer state; log is the per-token size of the recurrent-layer input log.}
\label{tab:r9-microbench}
\setlength{\tabcolsep}{3.2pt}
\resizebox{\columnwidth}{!}{%
\begin{tabular}{@{}lrrrrrr@{}}
\toprule
Model & Checkpoint & Log & \multicolumn{2}{c}{Replay $a+b\cdot L$} & Prefill & Leverage \\
\cmidrule(lr){4-5}
 & (MiB) & (KiB/tok) & $a$ (ms) & $b$ ($\mu$s/tok) & $c$ ($\mu$s/tok) & $c/b$ \\
\midrule
Qwen3.5-4B & 49.1 & 363 & 1.49 & 3.90 & 22.5 & 5.8$\times$ \\
Qwen3.5-9B & 49.1 & 363 & 1.48 & 3.89 & 53.7 & 13.8$\times$ \\
Qwen3.5-27B & 146.8 & 957 & 3.78 & 9.76 & 202.1 & 20.7$\times$ \\
\addlinespace[2pt]
Falcon-H1-7B & 66.9 & 312 & 1.02 & 3.00 & 33.5 & 11.2$\times$ \\
Falcon-H1-34B & 146.1 & 729 & 1.97 & 6.53 & 241.3 & 36.9$\times$ \\
\bottomrule
\end{tabular}%
}
\end{table}

%% file: tables/tab3-main-summary.tex
\begin{table*}[!tb]
\centering
\small
\caption{\textbf{\sys{} leads exact caches on goodput, tail latency, and 80\% SLO capacity.} SuffixReplay$^{*}$ (SR$^{*}$) is approximate and targets interleaved models.}
\label{tab:r10-summary}
\setlength{\tabcolsep}{4.2pt}
\resizebox{\textwidth}{!}{%
\begin{tabular}{@{}ll cccc cc ccc@{}}
\toprule
 & & \multicolumn{4}{c}{Anchor goodput / vLLM} & \multicolumn{2}{c}{Anchor latency (vLLM $\rightarrow$ \sys{})} & \multicolumn{3}{c}{80\% SLO capacity (calls/s)} \\
\cmidrule(lr){3-6}\cmidrule(lr){7-8}\cmidrule(l){9-11}
Model & Workload ($\lambda$) & \sys{} & SR$^{*}$ & Marconi & SPC & TTFT p50 (ms) & TPOT p99 (ms) & vLLM & \sys{} & Gain \\
\midrule
\multirow{4}{*}{Qwen3.5-9B}
 & ShareGPT ($\lambda{=}4$) & \textbf{1.27}{\tiny\,[1.21,\,1.32]} & 1.27 & 0.79 & 0.77 & 168 $\rightarrow$ \textbf{96} & 42 $\rightarrow$ \textbf{30} & 3.41 & \textbf{4.43} & \textbf{1.30$\times$} \\
 & SWE-agent ($\lambda{=}2$) & \textbf{1.11}{\tiny\,[1.09,\,1.12]} & 1.09 & 0.45 & 0.86 & 173 $\rightarrow$ \textbf{107} & 81 $\rightarrow$ \textbf{55} & 1.66 & \textbf{2.20} & \textbf{1.33$\times$} \\
 & LongBench ($\lambda{=}4$) & \textbf{1.23}{\tiny\,[1.22,\,1.24]} & 1.21 & 0.00 & 0.69 & 114 $\rightarrow$ \textbf{62} & 181 $\rightarrow$ \textbf{179} & 2.14 & \textbf{2.46} & \textbf{1.15$\times$} \\
 & ToT ($\lambda{=}8$) & \textbf{1.13}{\tiny\,[1.07,\,1.19]} & 0.89 & 0.67 & 0.85 & 80 $\rightarrow$ \textbf{68} & 75 $\rightarrow$ \textbf{41} & 10.25 & \textbf{12.86} & \textbf{1.25$\times$} \\
\midrule
\multirow{4}{*}{Falcon-H1-7B}
 & ShareGPT ($\lambda{=}4$) & \textbf{1.22}{\tiny\,[1.17,\,1.27]} & n/a & 1.15 & 0.77 & 667 $\rightarrow$ \textbf{246} & 44 $\rightarrow$ \textbf{36} & 2.72 & \textbf{2.86} & \textbf{1.05$\times$} \\
 & SWE-agent ($\lambda{=}2$) & \textbf{4.26}{\tiny\,[3.50,\,5.01]} & n/a & 2.39 & 0.82 & 1.1\,s $\rightarrow$ \textbf{125} & $\infty$ $\rightarrow$ \textbf{100} & 0.69 & \textbf{1.11} & \textbf{1.60$\times$} \\
 & LongBench ($\lambda{=}2$) & \textbf{1.13}{\tiny\,[1.11,\,1.16]} & n/a & 0.25 & 0.95 & 154 $\rightarrow$ \textbf{82} & 215 $\rightarrow$ \textbf{207} & 1.23 & \textbf{1.63} & \textbf{1.32$\times$} \\
 & ToT ($\lambda{=}8$) & \textbf{1.19}{\tiny\,[1.15,\,1.23]} & n/a & 0.66 & 0.77 & 129 $\rightarrow$ \textbf{98} & 127 $\rightarrow$ \textbf{48} & 4.40 & \textbf{7.60} & \textbf{1.73$\times$} \\
\bottomrule
\end{tabular}%
}
\end{table*}

%% file: tables/tab4-contested.tex
\begin{table}[!htb]
\centering
\caption{\textbf{Against approximate SuffixReplay$^{*}$, exact \sys{} ties on ShareGPT and SWE-agent and wins on LongBench and ToT.} Goodput ratio \sys{}\,/\,SuffixReplay$^{*}$ at the highest contested loads on Qwen3.5-9B, 32\,GiB.}
\label{tab:r9-contested}
\setlength{\tabcolsep}{3.5pt}
\resizebox{\columnwidth}{!}{%
\begin{tabular}{@{}lcccc l@{}}
\toprule
Workload & Run 1 & Run 2 & Run 3 & Mean [95\% CI] & Verdict \\
\midrule
ShareGPT ($\lambda{=}8$) & 0.978 & 1.023 & 0.956 & 0.986 [0.90, 1.07] & tie \\
SWE-agent ($\lambda{=}4$) & 1.020 & 1.012 & 1.006 & 1.013 [1.00, 1.03] & tie \\
SWE-agent ($\lambda{=}6$) & 1.054 & 1.065 & 0.974 & 1.031 [0.91, 1.15] & tie \\
LongBench ($\lambda{=}6$) & 1.550 & 1.551 & 1.513 & 1.538 [1.48, 1.59] & win \\
ToT ($\lambda{=}16$) & 1.091 & 1.130 & 1.082 & 1.101 [1.04, 1.16] & win \\
\bottomrule
\end{tabular}%
}
\end{table}